\documentclass[12pt, final]{l4dc2023}

\title[Hybrid Multi-agent DRL for AMoD]{Hybrid Multi-agent Deep Reinforcement Learning for\\ Autonomous Mobility on Demand Systems}

\usepackage{times, bm, mathtools, dsfont, etoolbox, comment}

\makeatletter
\patchcmd{\@algocf@start}% <cmd>
  {-1.5em}% <search>
  {0pt}% <replace>
  {}{}% <success><failure>
\makeatother

\author{%
 \Name{Tobias Enders} \Email{tobias.enders@tum.de}\\
 \addr Technical University of Munich, Germany
 \AND
 \Name{James Harrison} \Email{jamesharrison@google.com}\\
 \addr Google Research, Brain Team, California, USA
 \AND
 \Name{Marco Pavone} \Email{pavone@stanford.edu}\\
 \addr Stanford University, California, USA
 \AND
 \Name{Maximilian Schiffer} \Email{schiffer@tum.de}\\
 \addr Technical University of Munich, Germany
}

\begin{document}

\maketitle

\begin{abstract}
    We consider the sequential decision-making problem of making proactive request assignment and rejection decisions for a profit-maximizing operator of an autonomous mobility on demand system. We formalize this problem as a Markov decision process and propose a novel combination of multi-agent Soft Actor-Critic and weighted bipartite matching to obtain an anticipative control policy. Thereby, we factorize the operator's otherwise intractable action space, but still obtain a globally coordinated decision. Experiments based on real-world taxi data show that our method outperforms state of the art benchmarks with respect to performance, stability, and computational tractability. 
\end{abstract}


\section{Introduction}
Mobility on demand (MoD) systems, in which a fleet of free-floating vehicles serves customers' ad hoc requests for point-to-point transportation, have transformed urban mobility in recent years. Companies like Uber, Lyft, and DiDi, made MoD more accessible compared to taxi-based ride hailing services. Autonomous vehicles will further transform MoD systems; besides much lower prices, a major benefit of autonomous MoD (AMoD) is its improved potential for advanced control strategies, as a central operator obtains full control over the entire fleet. This transformation changes the fleet operator's control problem substantially: MoD operators focus primarily on revenue maximization, as human drivers' income is (almost) a fixed cost that dominates mileage-dependent operational cost. Contrarily, AMoD operators focus on the maximization of their operating profit, because operational costs dominate their total cost balance. In this context, the central operator can leverage its full knowledge about the system state and fleet control to make improved (proactive) dispatching decisions, i.e., request to vehicle assignment and rejection, to maximize its profit.

Since an operator does not have profound knowledge about future trip requests, it faces an online decision-making problem in a stochastic environment. Hence, it is promising to apply deep reinforcement learning (DRL) to this problem. However, AMoD systems entail many vehicles and trip requests, such that an operator's action space is very large and possibly time-varying as the number of requests changes over time. It is thus infeasible to apply off-the-shelf single-agent DRL. To solve this problem, we propose a novel combination of a multi-agent DRL algorithm with optimization-based centralized decision-making through weighted bipartite matching. This hybrid algorithm combines the advantages of multi-agent approaches, DRL, and combinatorial optimization. 

\subsection{Related Work}
To keep this literature overview concise, we focus on literature for controlling (autonomous) MoD systems in the following. For a review of multi-agent DRL, we refer to \cite{Gronauer2022} and further elaborate on how we build on the multi-agent DRL literature in Section~\ref{sec:method}. 

Classical approaches for dispatching and explicit rebalancing decisions focused on greedy or hand-crafted feature-based policies \citep{Liao2003,Zhang2017}, queueing theoretical approaches \citep{Zhang2016}, and model predictive control (MPC) \citep{AlonsoMora2017}.

Recently, many works applied DRL in the context of (autonomous) MoD, often including or purely focusing on explicit rebalancing \citep[e.g.,][]{Jiao2021, Gammelli2021, Skordilis2022, Liang2022}. Contrarily, other works focused on DRL for non-myopic dispatching, which entails an implicit rebalancing decision that avoids additional costs due to empty driving. Early approaches \citep[cf.][]{Xu2018, Wang2018} were shown to be inferior to at least one of the subsequent works: \cite{Li2019} proposed a mean field multi-agent actor-critic algorithm. \cite{Tang2019} used bipartite matching based on learned $V$-values. \cite{Zhou2019} combined a multi-agent Deep Q-Network with minimization of the Kullback-Leibler divergence (KL-divergence) between the vehicle and the request distribution. Finally, \cite{SadeghiEshkevari2022} described how DiDi recently rolled out DRL for dispatching in practice. 

Since these works strive to improve the operations of today's MoD systems, they aim at maximizing the drivers' revenue or the number of orders served, rather than at maximizing the profit of an AMoD system. While \cite{Xu2018, Wang2018, Tang2019, SadeghiEshkevari2022, Liang2022} also use a combination of multi-agent DRL and weighted matching, they all employ value-based algorithms. Contrarily, we use an actor-critic algorithm, enabling more advanced strategies to mitigate problems arising from multi-agent learning, in particular, decentralized actors with centralized critics, see Section~\ref{sec:method}. By using Soft Actor-Critic (SAC) \citep{Haarnoja2018}, we can nevertheless benefit from the improved sample-efficiency of off-policy algorithms.

\subsection{Contributions}
To the best of our knowledge, we are the first to consider the problem of making proactive dispatching decisions for a profit-maximizing AMoD system operator with DRL. We propose a novel method that combines multi-agent SAC with centralized final decision-making through weighted matching. We perform experiments based on real-world data and, similar to related works, benchmark our method against a greedy policy. In addition, we are the first to compare our method against an MPC approach. We show that our method outperforms the greedy policy on all instances by up to 5\%. Moreover, we outperform the MPC approach in most cases. Our DRL method shows a significantly more stable performance across varying instances, while MPC may perform arbitrarily bad---in single cases up to 60\% worse than the greedy policy. 
%An extended version of this paper, including an appendix, is available at \url{https://arxiv.org/abs/2212.07313}. 
Our code can be found at \url{https://github.com/tumBAIS/HybridMADRL-AMoD}.
%An extended version of this paper, including an extensive appendix to substantiate all results and (algorithmic) design decisions, is available at \url{https://arxiv.org/abs/2212.07313}. Our code can be found at \url{https://github.com/tumBAIS/HybridMADRL-AMoD}.

\section{Problem Formulation: Markov Decision Process}\label{sec:MDP}
We consider a profit-maximizing operator who centrally controls a fixed-size fleet of vehicles to serve customer trip requests revealed over time within an operating area. The operator can accept or reject requests and dispatches accepted requests to vehicles. These decisions must be made in real-time and immediately, i.e., the operator cannot defer requests to a later time step, as customers are not willing to wait for feedback. If the operator accepts a request, customers must be picked up within a known maximum waiting time $\omega^\text{max}\in\mathbb{N}_0$ after the request was placed. We formalize this control problem as a Markov decision process (MDP) as follows. 

\textbf{Preliminaries.} We consider a discrete time horizon $\mathcal T=\left\{0,1,...,T\right\}$. During one time step, multiple requests can enter the system. The operator makes one decision per time step for multiple requests simultaneously, which allows to optimize over a batch of requests. We represent the operating area as a graph $G=(V,E)$ with weight vectors $\prescript{e}{}{\bm w}=\left(\prescript{e}{}{w}^1, \prescript{e}{}{w}^2\right)\in\mathbb R_{>0}\times\mathbb N$, denoting the distance ($\prescript{e}{}{w}^1$) of and the time steps ($\prescript{e}{}{w}^2$) to traverse an edge $e\in E$. The nodes of $G$ may represent, e.g., the centers of zones into which the operating area is divided. 

\textbf{States.} We describe the system state at time $t\in\mathcal T$ by $\bm S_t=\bigl(t, \left(\prescript{t}{}{\bm r}^i\right)_{i\in\left\{1,...,R_t\right\}} \! , \bigl(\bm k_t^j\bigr)_{j\in\left\{1,...,K\right\}}\bigr)$, with $R_t$ being the variable number of new requests $\prescript{t}{}{\bm r}^i$, $i\in\{1,...,R_t\}$, at time step $t$, and $K$ vehicles $\bm k_t^j$, $j\in\{1,...,K\}$. A request $\bm r=\left(\omega,o,d\right)$ consists of a waiting time $\omega\in\mathbb N_0\cup\emptyset$, an origin $o\in V$, and a destination $d\in V\setminus \{o\}$; $\omega$ tracks the elapsed time from request placement to pickup, where we set $\omega\leftarrow\emptyset$ at pickup. We denote a vehicle by $\bm k = (v,\tau,\bm r^1,\bm r^2)$, with position $v\in V$ and the number of time steps $\tau\in\mathbb N_0$ left to reach this position. Here, $v$ can either be the current node if the vehicle idles or the next node that will be reached if the vehicle travels. Furthermore, slightly abusing notation, a vehicle can have at most two assigned requests $\bm r^1,\bm r^2$. Assigning more requests to one vehicle is unreasonable for realistic trip lengths and maximum waiting times. We denote the position of vehicle $\bm k_t^j$ by $\prescript{j \,}{}{\! v}_t$ and denote other components of the vehicle vector likewise.

\textbf{Actions.} The action space describing feasible decisions of the operator is
\begin{align}
    \mathcal A\left(\bm S_t\right)=\biggl\{ \left(a_t^1,...,a_t^{R_t}\right)\biggl\lvert\ & a_t^i=0\ \lor\ \left(a_t^i=j\in\left\{1,...,K\right\}\ \wedge\ \prescript{j}{}{\bm r}_t^2=\emptyset\right)\ \forall\ i\in\left\{1,...,R_t\right\}, \nonumber \\[-0.4em]
    & \sum_{i=1}^{R_t}\mathds 1\!\left(a_t^i=j\right)\leq 1\ \ \forall j\in\left\{1,...,K\right\} \biggr\} . \label{eq:action space}
\end{align}
The operator can take one decision $a_t^i$ per request $\prescript{t}{}{\bm r}^i$, $i\in\left\{1,...,R_t\right\}$, either rejecting it ($a_t^i=0$), which means that the request leaves the system, or assigning it to vehicle $\bm k^j$ ($a_t^i=j$), which is only possible if the vehicle does not already have two assigned requests, i.e., if $\prescript{j}{}{\bm r}_t^2=\emptyset$ holds. The final condition in (\ref{eq:action space}) implies that at most one new request is assigned to each vehicle in each time step, which is a realistic simplification facilitating the application of a matching algorithm. The central operator's action space size is of order $(K+1)^{R_t}$.

\textbf{Transitions.} We first describe the action-dependent transition from the pre-decision to post-decision state. Then, we describe the transition from the post-decision state to the next pre-decision state, which is independent of the action and only determined by the system dynamics. 

A reject decision has no impact on the state. When $a^i=j$, we add the request to the vehicle state, i.e., if $\prescript{j}{}{\bm r}^1=\emptyset$, then $\prescript{j}{}{\bm r}^1\leftarrow\prescript{t}{}{\bm r}^i$, and $\prescript{j}{}{\bm r}^2\leftarrow\prescript{t}{}{\bm r}^i$ otherwise. 

The following transitions apply to all vehicles: if the vehicle picks up a customer, i.e., if $\bm r^1\neq\emptyset\ \wedge\ \tau=0\ \wedge\ v=o\!\left(\bm r^1\right)$, where $o\!\left(\bm r^1\right)$ denotes the origin of request $\bm r^1$, then $\omega\!\left(\bm r^1\right)\leftarrow\emptyset$. If the vehicle moves between two nodes, i.e., if $\tau>0$, then $\tau\leftarrow\tau-1$. If the vehicle is at a node but moves to serve a request, i.e., if $\tau=0\ \wedge\ \bm r^1\neq\emptyset$, then $v$ is replaced by the next node $v'$ on the vehicle's route to serve the request, going to origin or from origin to destination, and $\tau\leftarrow\prescript{(v,v')}{}{w}^2-1$. If a vehicle drops off a customer before the next decision is made, i.e., if $\bm r^1\neq\emptyset\ \wedge\ \omega\!\left(\bm r^1\right)=\emptyset\ \wedge\ \tau=0\ \wedge\ v=d\!\left(\bm r^1\right)$, we shift requests: $\bm r^1\leftarrow\bm r^2$ and $\bm r^2\leftarrow\emptyset$. We increment the waiting times $\omega\neq\emptyset$ of requests that have not been picked up yet, i.e., $\omega\leftarrow\omega+1$, where $\omega$ refers to $\omega\!\left(\bm r^1\right)$ and/or $\omega\!\left(\bm r^2\right)$. Moreover, independent of the vehicles' states, customers place new requests, i.e., $\left(\prescript{t}{}{\bm r}^i\right)_{i\in\left\{1,...,R_t\right\}}$ is replaced by $\left(\prescript{t+1}{}{\bm r}^i\right)_{i\in\left\{1,...,R_{t+1}\right\}}$. Note that we do not know the underlying time-dependent probability distribution which generates new requests, but we can simulate the resulting requests by replaying historic data. We assume that the new requests arrive independently of the state-action history, such that the Markov property holds. Finally, $t\leftarrow t+1$.

\textbf{Rewards.} Since the operator maximizes its profit and fixed costs are independent of the control problem, our reward function focuses on the operating profit, which is the revenue from serving requests minus operational costs, e.g., for fuel and maintenance. The operator obtains the revenue for a request $\bm{r}$ when a vehicle picks up the request within the maximum waiting time. The revenue is given by a function $\mathrm{rev}(\bm r)\in\mathbb R_{>0}$, representing the operator's pricing model. For improved readability, we express the profit components as functions of the post-decision state $\bm S_{t^+}$ and write $t$ for $t^+$. Then, the total revenue at time $t$ is
\begin{equation*}
    \mathrm{Rev}(\bm S_t)=\sum_{j=1}^K \, \mathds 1\Bigl( \prescript{j}{}{\bm r}_t^1\neq\emptyset\ \wedge\ \prescript{j}{}{\tau}_t=0\ \wedge\ \prescript{j}{}{v}_t=o\!\left(\prescript{j}{}{\bm r}_t^1\right)\ \wedge\ \omega\!\left(\prescript{j}{}{\bm r}_t^1\right)\leq\omega^\text{max} \Bigr)\cdot \mathrm{rev}\!\left(\prescript{j}{}{\bm r}_t^1\right).
\end{equation*}
When a vehicle starts to move from $v$ to $v'$, the operator incurs operational costs $c\in\mathbb R_{>0}$ per distance unit, as commonly assumed \citep[see, e.g.,][]{Boesch2018}. Thus, the total cost at time $t$ is
\begin{equation*}
    \mathrm{Cost}(\bm S_t)=c\cdot\sum_{j=1}^K \, \mathds 1 \Bigl(\prescript{j}{}{\tau}_t=0\ \wedge\ \prescript{j}{}{\bm r}_t^1\neq\emptyset\Bigr)\cdot\prescript{\left(\prescript{j}{}{v}_t,\ \prescript{j}{}{v}_t'\right)}{}{w}^1 .
\end{equation*}
The total profit at time $t^+$ is $\mathrm{Profit}(\bm S_{t^+})=\mathrm{Rev}(\bm S_{t^+})-\mathrm{Cost}(\bm S_{t^+})$. Note that $\bm S_{t^+}$ is a function of $\bm S_t$ (pre-decision) and $\bm a_t\in\mathcal A\left(\bm S_t\right)$, such that we write $\mathrm{Profit}(\bm S_{t^+})=\mathrm{Profit}(\bm S_t, \bm a_t)$.

The AMoD operator wants to find a policy $\pi\left(\bm a_t\lvert\bm S_t\right)$ that maximizes the expected total reward over all time steps, given the initial state $\bm S_0$:
\begin{equation*}
    \mathrm{Profit}^\ast\!\left(\bm S_0\right)=\max_\pi\, \mathbb E_{\left(\bm S_t, \bm a_t\right)\sim\pi}\left[\, \left. \sum_{t=0}^{T-1}\mathrm{Profit}\left(\bm S_t, \bm a_t\right)\, \right\rvert\, \bm S_0\, \right] .
\end{equation*}
To do so, we propose a hybrid DRL algorithm in the following section. 

\section{Method: Multi-agent Soft Actor-Critic with Global Matching}\label{sec:method}
Analyzing our problem setting, we identify two key requirements to develop an algorithm that constructs an effective control policy: first, it should leverage information patterns that can be observed from historic trip data to make non-myopic decisions. Second, it should be scalable to a realistic system size to coordinate a large number of vehicles and requests. To account for the second requirement, we formalize the centralized dispatching of vehicles to requests as a bipartite matching problem (BMP). This BMP should be weighted to allow for non-myopic dispatching decisions, anticipating the downstream impact of decisions in a stochastic environment. The choice of weights heavily impacts the policy's performance. Therefore, we use DRL to parameterize these weights, as it accounts for the downstream impact of decisions in stochastic environments by design and allows to extract and use information from historic data---thus, covering the first requirement. 

However, single-agent DRL is not suitable for our problem setting, as the central operator's action space scales exponentially with the number of vehicles and requests per time step and becomes intractable very quickly. Thus, we leverage multi-agent DRL to factorize the action space at the price of increased complexity, caused by having to coordinate the actions of multiple DRL agents to finally take a centralized decision. Our hybrid algorithm combines the advantages of multi-agent DRL with those of combinatorial optimization: we use DRL agents as estimators to compute non-myopic weights, serving as the input to a weighted bipartite matching algorithm, which then makes a globally optimal and coordinated decision. 

\subsection{Overview}\label{sec:method_overview}
\begin{figure}
    \includegraphics[width=\textwidth]{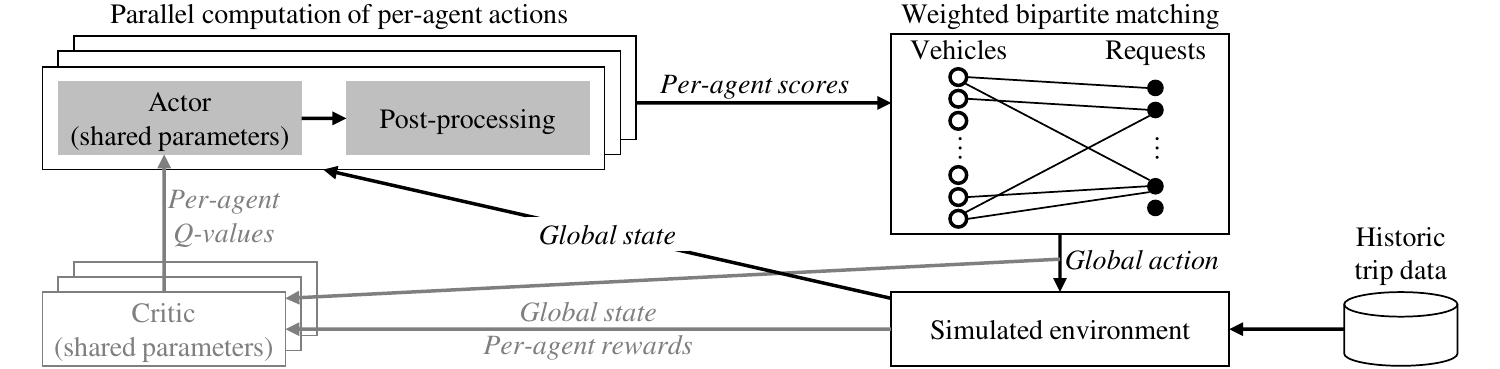}
    \caption{Overview of our method (gray text refers to parts which we use only during training).}
    \label{fig:methodology_overview}
\end{figure}

Figure~\ref{fig:methodology_overview} provides an overview of our method in which we leverage a DRL algorithm to parameterize a weighted bipartite matching to take anticipatory global dispatching decisions. To obtain a weight for each request-vehicle combination, we consider each combination as one agent. We represent these agents by an actor network, which we train using the SAC algorithm \citep{Haarnoja2018}. To obtain the weights for our BMP, we post-process the actors' outputs, such that from the perspective of the DRL agents and the computation of policy parameter gradients, post-processing and matching are part of the environment.

Since the DRL agents are a means to factorize the action space of the central operator, rather than ``real" individual agents, they can observe the global system state. Although they should take cooperative decisions that eventually benefit the central operator's profit, the agents observe their own (egoistic) rewards, not the global system reward, to avoid a credit assignment problem \citep[e.g.,][]{Agogino2004} that otherwise occurs, in particular with many agents. Here, we enforce coordination of the agents through the BMP and note that varying the credit assignment scheme remains an interesting question for future work. 

Individual rewards imply a need for per-agent critic values. From one agent's perspective, the other agents are part of the environment. Since we train all agents concurrently, their policies change simultaneously and the perceived environment is non-stationary. To mitigate this, the critic gets the other agents' actions as an additional input (``centralized critic"), such that the policy evaluation can explicitly account for other agents' behavior \citep[e.g.,][]{Lowe2017, Iqbal2019}. 

All agents represent a request-vehicle combination and are thus homogeneous. Accordingly, they can share parameters and we need only one actor and one critic network for all agents \citep[cf.][]{Iqbal2019}. We can train those centrally, i.e., the total parameter update is given by the sum of per-agent updates. Still, the forward pass of the actor network is independent across agents, allowing for decentralized and parallelized execution, which is important for scalability. 

Our method can handle a variable number of requests and thus a variable global action space size, as we can use neural networks with parameter sharing for any number of agents in parallel and the BMP does not require a fixed number of requests. The same holds true for the vehicles, such that the system size when testing may differ from training (see appendix). 

We provide details on the individual components of our method in the following.

\subsection{Per-agent Post-processing and Global Matching}
The actor network parameterizes a categorical probability distribution over the two actions that can be taken for a request-vehicle combination: reject or accept. We post-process the actor output per agent to transform it into a per-agent score, that we then use in the global weighted matching.

Algorithm~\ref{alg:post-processing} defines the post-processing. First, we mask infeasible actions by setting the accept probability $p_\text{a}$ to zero if the vehicle already has two assigned requests. Then, we sample a reject/assign decision from the masked probability distribution; when testing, we instead take the argmax of the probabilities. A reject decision ($\delta=0$) at the per-agent level implies a request-to-vehicle reject decision at the global level, such that we set the respective score to zero. For an accept decision, we use the accept probability as score\footnote{At first sight, it might seem more intuitive to let the actor parameterize a continuous distribution, from which we can take a sample to directly obtain the score, and/or have separate outputs for the reject/accept probability and the score. We tested both approaches, but empirically observed that they perform worse than the variant described here. If we choose one of those approaches, we cannot compute the terms in the loss functions (see Section~\ref{sec:loss_functions}) which are an expectation w.r.t. the policy, i.e., $\pi_\phi(a\,\lvert\, s,i)^T\cdot(...)$, since the action space is not (purely) categorical anymore. Then, we need to sample to estimate the expectation, as in the version of SAC for continuous action spaces, which increases the variance. We hypothesize that this harms the algorithm's performance and explains our empirical observation.}, such that a higher probability leads to a higher score for the weighted matching. An accept decision at the per-agent level does not always imply an accept decision at the global level, as the matching might assign the request to a different vehicle. 

\begin{algorithm2e}
    \caption{Per-agent post-processing}
    \label{alg:post-processing}
    \DontPrintSemicolon
    \KwIn{$p_\text{r}, p_\text{a}\in[0,1] \text{ s. t. } p_\text{r}+p_\text{a}=1 \ \ ; \ \ \bm k^j$}
    \KwOut{score $s$}
    %\vspace{0.2cm}
    \lIf(\tcp*[f]{\small reject if already two assigned requests}){$\prescript{j}{}{\bm r}^2\neq\emptyset$}
        {$p_\text{r}\gets 1\ ,\ p_\text{a}\gets 0$}
    \leIf{training}
        {$\delta\sim\mathrm{Categorical}\bigl((p_\text{r},p_\text{a})\bigr)$ \tcp*{\small sample $\delta\in\{0,1\}$ when training}}
        {$\delta\gets 0\text{ if } p_\text{r}\geq p_\text{a},\ \delta\gets1\text{ if } p_\text{r}< p_\text{a}$ \tcp*[f]{\small argmax when testing\hspace{-0.1cm}}}
    \leIf{$\delta=0$}
        {$s\gets 0$ \tcp*{\small score is zero if rejected}}
        {$s\gets p_\text{a}$ \tcp*[f]{\small score is accept probability if accepted\hspace{-0.1cm}}}
\end{algorithm2e}

We use all agents' scores to create a bipartite graph, with vehicles and requests as nodes, and edges between all vehicle and request nodes for which we obtain per-agent accept decisions (i.e., $s>0$). The edges' weights correspond to the respective scores. We solve the resulting maximum weighted BMP (formally defined in the appendix) using the Hungarian algorithm \citep{Kuhn1955} to get a globally coordinated decision, where each request is assigned at most once.

\subsection{Multi-agent Soft Actor-Critic}\label{sec:loss_functions}
SAC is an entropy-regularized, off-policy actor-critic algorithm. It trains a stochastic policy $\pi\left(\bm a_t\lvert\bm S_t\right)$ with entropy maximization, incentivizing exploration through a random policy:
\begin{equation*}
    \pi^\ast=\arg\max_\pi\, \mathbb E_{\left(\bm S_t, \bm a_t\right)\sim\pi}\left[\, \sum_{t=0}^{T-1} \mathrm{Profit}\left(\bm S_t,\bm a_t\right) + \alpha\, H\!\left(\pi\left(\, \cdot\, \lvert\, \bm S_t\right)\right) \right] .
\end{equation*}
The entropy of the policy is defined as $H\!\left(\pi\left(\, \cdot\, \lvert\, \bm S_t\right)\right)=-\mathbb E_{\bm a_t\sim\pi}\log\pi\left(\bm a_t\lvert\bm S_t\right)$ and the entropy coefficient $\alpha\in\mathbb R_{\geq 0}$ is a hyperparameter that controls the exploitation/exploration trade-off. While SAC was originally developed for continuous action spaces in \cite{Haarnoja2018}, it can also be applied to discrete actions \citep{Christodoulou2019}. 

We chose to use an actor-critic algorithm to enable our multi-agent approach of decentralized actors with centralized critics as explained in Section~\ref{sec:method_overview}. Since exploration is paramount for our problem setting, we use SAC, which lets us explicitly tune how much the policy explores. 

We parameterize the actor network with parameters $\phi$ and the critic with $\theta$. SAC uses two critic networks, $Q\in\left\{Q^1,Q^2\right\}$, as well as corresponding target networks with parameters $\bar\theta$, which are an exponential moving average of the primary parameters $\theta$. Based on the multi-agent approach as described in Section~\ref{sec:method_overview}, the loss function for the actor with shared parameters is
\begin{equation*}
    J_\pi(\phi)=\mathbb E_{s\sim D}\left[ \sum_i \pi_\phi(a\,\lvert\, s,i)^T\cdot \left( \alpha\log\pi_\phi(a\,\lvert\, s,i) - \min_{j\in\{1,2\}}\left\{ Q_\theta^j\left( a\,\lvert\, s,i,\bar a_{-i} \right) \right\} \right) \right] .
\end{equation*}
Here, we use a simplified notation for improved readability: we denote a transition by $\left(s,\bar a,r,s'\right)$, with global states $s,s'$, global action $\bar a$ (after the matching), and rewards $r$. For agent $i$, $r_i$ denotes its reward, $\bar a_{-i}$ is the global action except for agent $i$'s action, and $a$ is a per-agent action (reject/assign), such that $\pi_\phi(a\,\lvert\, s,i)\in[0,1]^2$ and $Q_\theta^j\left( a\,\lvert\, s,i,\bar a_{-i} \right)\in\mathbb R^2$. We sample states (or transitions) from the replay buffer $D$ and denote the discount factor by $\gamma$. For the actor loss, we do not sample the global action from the replay buffer, but compute it based on the state $s$ and the current policy, as in \cite{Iqbal2019}. For each of the two critics $Q\in\left\{Q^1,Q^2\right\}$, the loss function is
\begin{align*}
    J_Q(\theta)&=\mathbb E_{(s, \bar a, r, s')\sim D}\left[ \sum_i\frac12 \left( Q_\theta\left(a\,\lvert\, s,i,\bar a_{-i}\right)\Big\rvert_{\bar a_i}-y_i \right)^2 \right] \text{, with} \\
    y_i&=r_i+\gamma\cdot\pi_\phi\left(a'\,\big\lvert\, s', i\right)^T\cdot\left( \min_{j\in\{1,2\}} \left\{ Q_{\bar\theta}^j\left( a'\,\big\lvert\, s',i,\bar a'_{-i} \right) \right\}-\alpha\log\pi_\phi\left(a'\,\big\lvert\, s',i\right) \right) .
\end{align*}
Here, the notation $\big\rvert_{\bar a_i}$ means ``evaluated at $\bar a_i$", i.e., of the two $Q$-values that we compute for the two possible actions of agent $i$, we use the one corresponding to the global decision $\bar a$. The term after $\gamma$ is the $V$-value estimate for $s'$ based on $\bar\theta$, for which we compute the next global action $\bar a'$ with the current policy. The number of requests and thus the number of agents can change between subsequent time steps. This poses a numerical problem for the critic loss computation, which requires the same number of agents for $s$ and $s'$. We solve this problem by amending the requests in $s'$ when saving a transition to the replay buffer and provide details on this in the appendix. 

The actor network obtains all vehicle states and requests for which a decision must be made in the current time step as an input. To deal with these (potentially) many inputs, we train a single request embedding and a single vehicle embedding to encode all requests and vehicles, respectively. To account for the variable number of requests and to let each agent focus on the parts of the input that are important for this particular agent, we equip the neural network with an attention mechanism \citep[cf.][]{Holler2019, Kullman2022}. Together with the request and vehicle embeddings for the agent and additional features, we pass the context computed by the attention mechanism to a sequence of feedforward layers. The critic network has the same architecture, but receives the global action as an additional input. We remove the action of the agent from this input, since the critic outputs $Q$-values for both possible actions. Further details on the neural networks, e.g., a formal description of the attention mechanism and hyperparameters, can be found in the appendix.

\section{Experiments}\label{sec:experiments}
To validate our method, we perform experiments based on historic taxi data that is publicly available for New York City \citep{NYCTLC2015}. We use a hexagon grid for spatial discretization and consider two different instances: one with 11 small zones (approx. 500 meters distance between neighboring zones) and one with 38 large zones (approx. 1 km distance), both in Manhattan. We consider the time interval from 8:30 am to 9:30 am during morning rush hour as one episode. Our data set contains data for 245 different dates in 2015, which we split into 200 training dates, 25 validation dates, and 20 test dates. We use a time step size of one minute and choose revenue and cost parameters such that a vehicle that serves a customer without empty driving achieves an operating profit margin of 10\%. We consider different numbers of vehicles to simulate different degrees of supply shortage; only cases with supply shortage are interesting, as the operator can serve each request immediately in the case of infinite supply, such that a myopic policy would be sufficient. For additional details on the data set, system setup, and hyperparameters, we refer to the appendix.

To benchmark our method, we compare its test performance against two ``classical", non RL-based, algorithms: a greedy policy and an MPC approach. The greedy policy accepts any request that can be served with a positive profit (accounting for the cost for empty driving to the pickup location) and rejects all others. It is a reasonable choice when there is no reliable estimate of future requests. If we have such an estimate, it is promising to apply MPC \citep[see, e.g.,][]{AlonsoMora2017}. We adapt this approach to our setting, using a request distribution estimate for mixed-integer-based receding-horizon optimization. Details on both benchmarks can be found in the appendix.

\section{Results and Discussion}\label{sec:results}
We provide plots illustrating the training process in the appendix. Figure~\ref{fig:box_plots} summarizes the performance of greedy, MPC, and our RL method on the test data for all considered instances. On average, our RL method always outperforms the greedy policy, by up to 5\% over the 20 test dates. For individual dates, RL outperforms greedy by up to 17\%. It performs by at most 6\% worse than greedy for less than 20\% of the individual dates. MPC is (substantially) worse than greedy and RL in many cases, although it sometimes outperforms the RL method. This means that MPC can provide a benefit in certain situations, but comes with an unstable performance across instances, which limits its practical applicability. In particular, MPC does not perform well in situations where there is a large shortage of vehicles, which are handled well by our RL method. Thus, our method provides a stable alternative, that always achieves at least the greedy performance and outperforms it by a substantial margin in many cases. Note that the order of magnitude of this performance improvement is significant for our application area \citep[cf.][]{SadeghiEshkevari2022}. Given the large scale at which AMoD systems operate, the seemingly small percentage improvements translate into significant monetary value for the operator. 
\begin{figure}
    \centering
    \includegraphics[width=\textwidth]{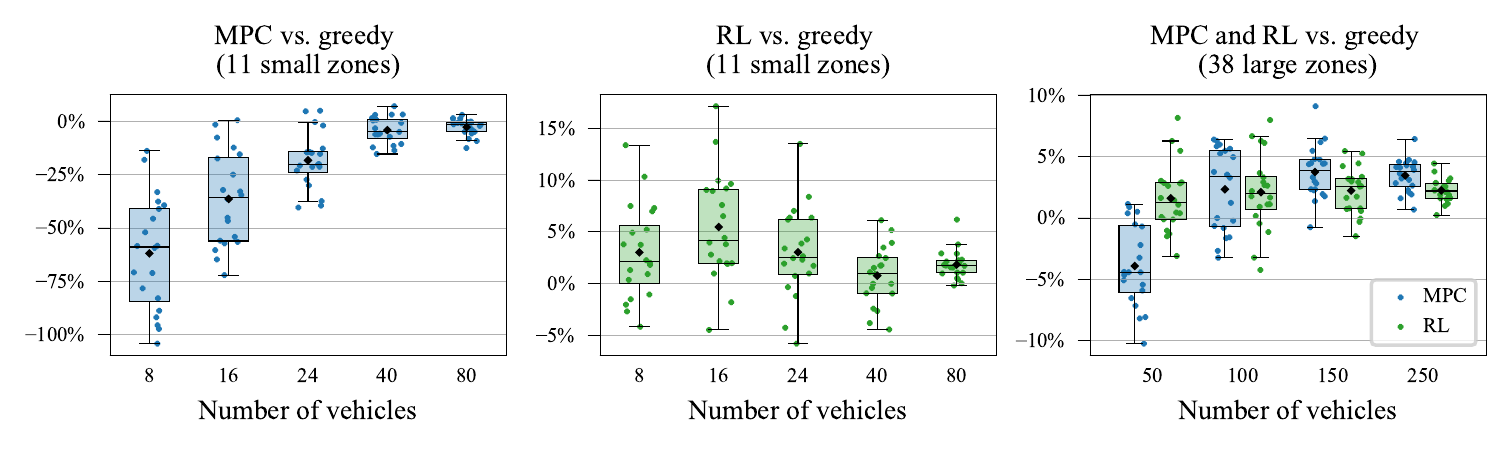}
    \caption{Test performance of MPC and our RL method compared to greedy (greedy is 0\%, values $<0\%$ indicate a performance worse than greedy). Each dot represents one test date.}
    \label{fig:box_plots}
\end{figure}

\begin{figure}
    \centering
    \includegraphics[width=\textwidth]{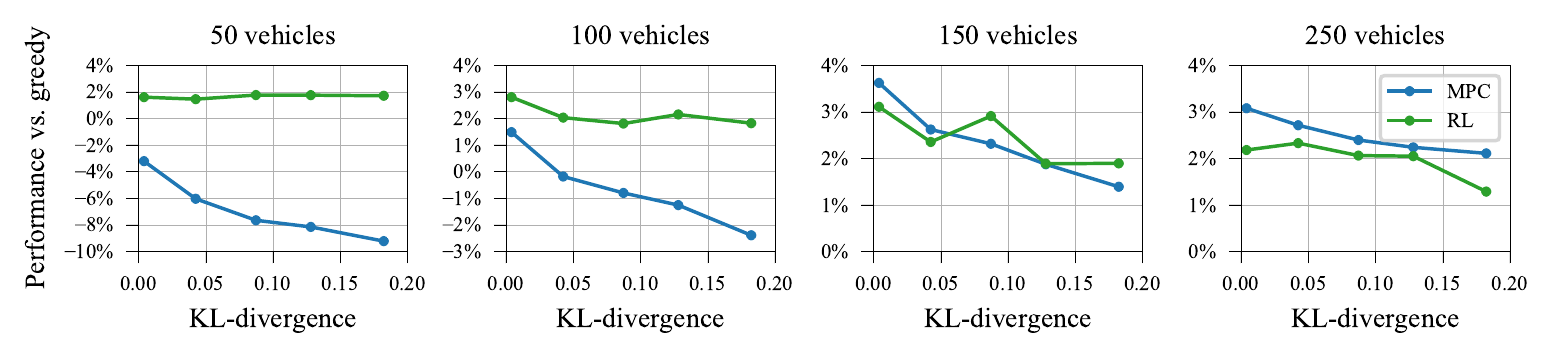}
    \caption{Mean test performance of MPC and RL vs. greedy as a function of the KL-divergence between the true request distribution and the distribution estimated from the training data (with different amounts of training data, resulting in different KL-divergence values). We use the request distribution estimated from the real data for 38 large zones as the true distribution and run the experiments with synthetic data sampled from this distribution.}
    \label{fig:kl_divergence}
\end{figure}

Figure~\ref{fig:kl_divergence} shows the performance of MPC and RL for the instance with 38 large zones and different amounts of training data, i.e, different estimation qualities of the request probability distribution. Since our problem setting excludes fixed costs, additional resources, i.e., vehicles, are free of charge. Thus, the greedy policy performs better with very few or many vehicles, compared to instances with a medium number of vehicles. With few vehicles, many requests are available for each vehicle, such that vehicles are rarely idle or drive without a customer. With many vehicles, most requests can be served quickly without empty driving because vehicles are usually available. Consequently, with sufficient training data, for both MPC and our RL approach, the performance gain vs. greedy is largest for a medium number of vehicles. However, with few vehicles, MPC performs worse than greedy, as it is not robust against mistakes when sampling future requests. Such errors have a larger effect with fewer vehicles. With less training data, the performance gain of our method decreases by about one percentage point, but it remains reliably better than greedy. For all instances except the 250 vehicles case, the performance loss is much larger for MPC; it is not always able to sustain a performance better than greedy, even for instances where it outperforms greedy with more training data. With 250 vehicles, there are many resources free of charge, such that the mistakes made by MPC have such a small effect that it is robust against a poor estimation quality. Based on these observations, we conclude that our RL method is more robust against a poor estimation quality due to insufficient training data than MPC. These results might seem surprising, as RL is in general not very sample-efficient---although SAC has better sample-efficiency than most policy gradient-based algorithms, since it is an off-policy algorithm and uses a replay buffer. However, for our problem setting, less training data does not mean that the RL agents must learn from fewer samples, as the available training data can be replayed multiple times in the simulated environment. The performance loss that we observe for the RL method is more likely due to the decreasing diversity of the training data to which the RL agents are exposed, leading to less generalization.

\begin{figure}
    \centering
    \includegraphics{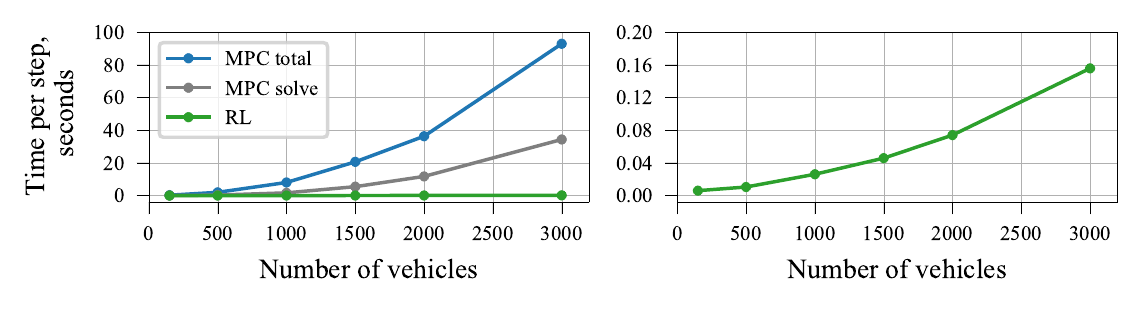}
    \caption{Time to compute one action (mean over 60 steps, based on 38 large zones, number of requests scaled with number of vehicles). MPC solve refers to solving the mixed integer program (MIP), MPC total also includes the generation of the MIP instance in each step.}
    \label{fig:runtime}
\end{figure}

Finally, a major advantage of our RL method over MPC is its shorter computational time during execution. We can train the network parameters offline in advance and easily scale the online execution, because the per-agent actor computations are fast and straightforward to parallelize. Figure~\ref{fig:runtime} shows that although we solve a combinatorial optimization problem in each time step, the computational time of our method is very short even for large system sizes. On the other hand, the computational time of MPC increases quickly for large instances: for 3000 vehicles, the action computation for a single step takes more than 30 seconds with MPC, while our RL method (including the matching) takes less than 0.2 seconds. Thus, the practical application of MPC at scale is greatly limited by its computational time, while our RL method can be scaled to much larger system sizes. 

\section{Conclusion}\label{sec:conclusion}
We consider the dispatching problem of a profit-maximizing AMoD operator with centralized control over a fleet of autonomous vehicles, who accepts (and serves) or proactively rejects requests in real-time. To solve this problem, we use a combination of multi-agent SAC with centralized final decision-making through weighted matching. Our experiments based on real-world data show that our method outperforms two strong benchmarks on most problem instances, that it is stable across these instances and robust against a poor estimation of the request distribution, and that it can be easily scaled to large system sizes. In future work, we will investigate the use of global instead of ego rewards. Furthermore, we will extend our framework to more complex use cases, e.g., dispatching and charge scheduling.

\newpage
\appendix

\section{Method}
In the following, we provide complementary details on our method. First, we describe how we assign per-agent rewards. Second, we formally define the weighted BMP. Third, we explain our solution to the dimensionality problem in the critic loss computation. Fourth, we provide additional details on the neural networks. Fifth, we give an overview of alternative approaches to post-processing and weighted matching which we tested and discarded. 

\subsection{Per-agent Rewards} 
Agents observe their own (egoistic) rewards. When matching agents, we can already compute the (potentially negative) profit that will result from the decision to assign a certain request to a certain vehicle. This profit consists of the revenue that will be obtained for the request, minus the operational costs to drive from the vehicle's position after it finished serving its current request (if any) to the new request's origin and the operational costs to drive from there to the new request's destination. If a request is matched to a vehicle at the global level, the corresponding agent immediately observes this profit as the reward. All agents for which the request is not matched to the vehicle at the global level observe a reward of zero. The sum over all these per-agent rewards equals the system reward, although we virtually forward rewards in time. When we sample transitions from the replay buffer, we normalize the sampled rewards by dividing them by the standard deviation of all rewards currently stored in the replay buffer \citep[cf.][]{Kurin2022}.

\subsection{Bipartite Matching Problem}
At time step $t$, the weighted BMP is formally defined as
\begin{align*}
    \max_{x_{ij}}\ &\sum_{i=1}^{R_t}\sum_{j=1}^K s_{ij}\cdot x_{ij} \\
    \text{s. t. } &\sum_{i=1}^{R_t} x_{ij} \leq 1\quad \text{for all}\ \ j\in\left\{1, ... , K\right\} , \\
    &\sum_{j=1}^K x_{ij} \leq 1\quad \text{for all}\ \ i\in\left\{1, ... , R_t\right\} ,
\end{align*}
where $s_{ij}$ is the score computed for request $\prescript{t}{}{\bm r}^i$ and vehicle $\bm k^j$, and the decision variables are $x_{ij}=1$ if $\prescript{t}{}{\bm r}^i$ is assigned to $\bm k^j$ and $x_{ij}=0$ otherwise.

\subsection{Critic Loss Computation Despite Variable Number of Agents}
The number of agents changes between time steps, since the number of requests varies over time. However, for the TD-update of the $Q$-function, we need the same number of agents at time $t$ and $t+1$. Thus, when adding transitions to the replay buffer, we replace the requests state for $t+1$ by the requests state for $t$ to obtain a matching number of agents. Given the small time step size, it is strongly stochastic which requests occur at which time step, and the underlying distribution can be assumed to be very similar for two consecutive time steps, such that the requests at $t$ are in expectation a similarly realistic sample for $t+1$ as the requests which we actually observe.

\subsection{Neural Networks}
We first describe the features which we use as inputs to the neural networks. We identify the zones which we use for spatial discretization by a horizontal and a vertical index. All locations (e.g., the request origin) correspond to one of these zones. We encode them by a vector with the horizontal and the vertical index, each normalized to $[0,1]$. We also tried to use sinusoidal positional encodings \citep[cf.][]{Vaswani2017} instead, but this did not improve the performance of our method. A request encoding then consists of:
\begin{itemize}
    \item The encoding of the request origin
    \item The encoding of the request destination
    \item The distance from origin to destination on the graph $G$, normalized to $[0,1]$
\end{itemize}
We encode a vehicle state by:
\begin{itemize}
    \item The encoding of the vehicle's position, where we use the current position of the vehicle if it does not have an assigned request or the destination of the assigned request that will be served last
    \item The number of time steps left to reach this position, normalized to $[0,1]$
    \item The number of assigned requests, normalized to $[0,1]$
\end{itemize}
Apart from request and vehicle states, we use some additional features: 
\begin{itemize}
    \item The current time step, normalized to $[0,1]$
    \item A flag in $\{0,1\}$ indicating if the vehicle under consideration will be able to serve the request under consideration within the maximum waiting time if it is matched to the vehicle
    \item The time steps to reach the position summed over all vehicles, normalized to $[0,1]$, indicating how busy the fleet currently is
    \item The number of requests placed since the current episode started, divided by the count of requests that are placed on average until the current time step, indicating how much demand was observed compared to an average episode
\end{itemize} 

Next, we describe the actor network. The request and vehicle encodings are used as the input for the request and vehicle embeddings, respectively. Both the request and the vehicle embedding are a feedforward layer with 32 units and ReLU activation. We denote the resulting embedding vectors by $\bm e_{r^i}$, $i\in\left\{1, ... , R_t\right\}$, and $\bm e_{k^j}$, $j\in\left\{1, ... , K\right\}$. The attention mechanism computes a global context, which is the concatenation of a requests context and a vehicles context. The requests context is computed as $\bm c_r=\sum_{i=1}^{R_t}\beta_{r^i}\cdot\bm e_{r^i}$ with $\beta_{r^i}=\sigma\left(\bm w_r\cdot\tanh\left(W_r\cdot\bm e_{r^i}\right)\right)\in\mathbb R$, where $\sigma$ is the sigmoid activation function and $\bm w_r\in\mathbb R^{256}$ as well as $W_r\in\mathbb R^{256\times32}$ are trainable parameters (the parameters are the same across all requests). The vehicles context is computed as $\bm c_k=\sum_{i=1}^{K}\beta_{k^i}\cdot\bm e_{k^i}$ with $\beta_{k^i}=\sigma\left(\bm w_k\cdot\tanh\left(W_k\cdot\bm e_{k^i}\right)\right)\in\mathbb R$ and trainable parameters $\bm w_k\in\mathbb R^{128}$, $W_k\in\mathbb R^{128\times32}$ (the parameters are the same across all vehicles). Thereby, we obtain a fixed size global representation of variable sized inputs. The structure of the embeddings and the attention mechanism is similar to \cite{Holler2019, Kullman2022}. We use the global context as well as the request and vehicle embedding corresponding to the agent under consideration together with the aforementioned additional inputs as the input to a sequence of feedforward layers. We use five layers with ReLU activation and 1024, 512, 128, 32, and 8 units, respectively, followed by the output layer of size 2 with softmax activation. For all layers, we use L2 regularization with a regularization coefficient of $10^{-4}$. We chose the specific architecture (number and size of layers) through hyperparameter tuning. 

The critic networks are identical to the actor network except for the following differences: For all request and vehicle states that do not correspond to the agent under consideration, we add the action to their encodings. For the requests, this is a flag in $\{0,1\}$ indicating if the request was rejected or accepted. For the vehicles, this is the origin and destination of the request that was newly assigned to the vehicle (zeros if no request was newly assigned to the vehicle). Moreover, we use the request and vehicle states of the agent under consideration as inputs to the feedforward layer of size 1024, but not for the embedding and attention layers. Finally, there is no activation function in the output layer.

\subsection{Alternatives to Post-processing and Weighted Matching}
We tested two alternatives to our weighted matching approach to obtain a global decision from the per-agent actor network outputs: generating a global probability distribution and non-weighted matching. 

For the global probability distribution, we use the per-agent reject/accept probabilities to construct the joined probability distribution over all feasible global decisions. We then sample, by taking the argmax when testing, from the global distribution to obtain a global decision. While this approach removes the need for a combinatorial optimization algorithm and gives promising results on very small instances, it becomes intractable quickly due to the exponentially increasing size of the global action space. 

For the non-weighted matching, we post-process the per-agent actor network outputs as described in Algorithm~\ref{alg:post-processing}, but do not use scores to construct a weighted bipartite graph. Instead, we create a non-weighted bipartite graph based on the per-agent reject/assign decisions $\delta$ and use a maximum (non-weighted) bipartite matching algorithm to obtain a global decision. While this approach is scalable, it performs worse than the weighted matching variant. 

\section{Experiments}
In the following, we provide complementary information on our experiments. First, we give more details on the system setup and how we pre-process the real-world taxi data set. Second, we state the hyperparameter values used in our experiments. Third, we provide details on the two benchmark algorithms. 

\subsection{Data Set and System Setup}
\begin{figure}[ht]
    \centering
    \includegraphics[width=\textwidth]{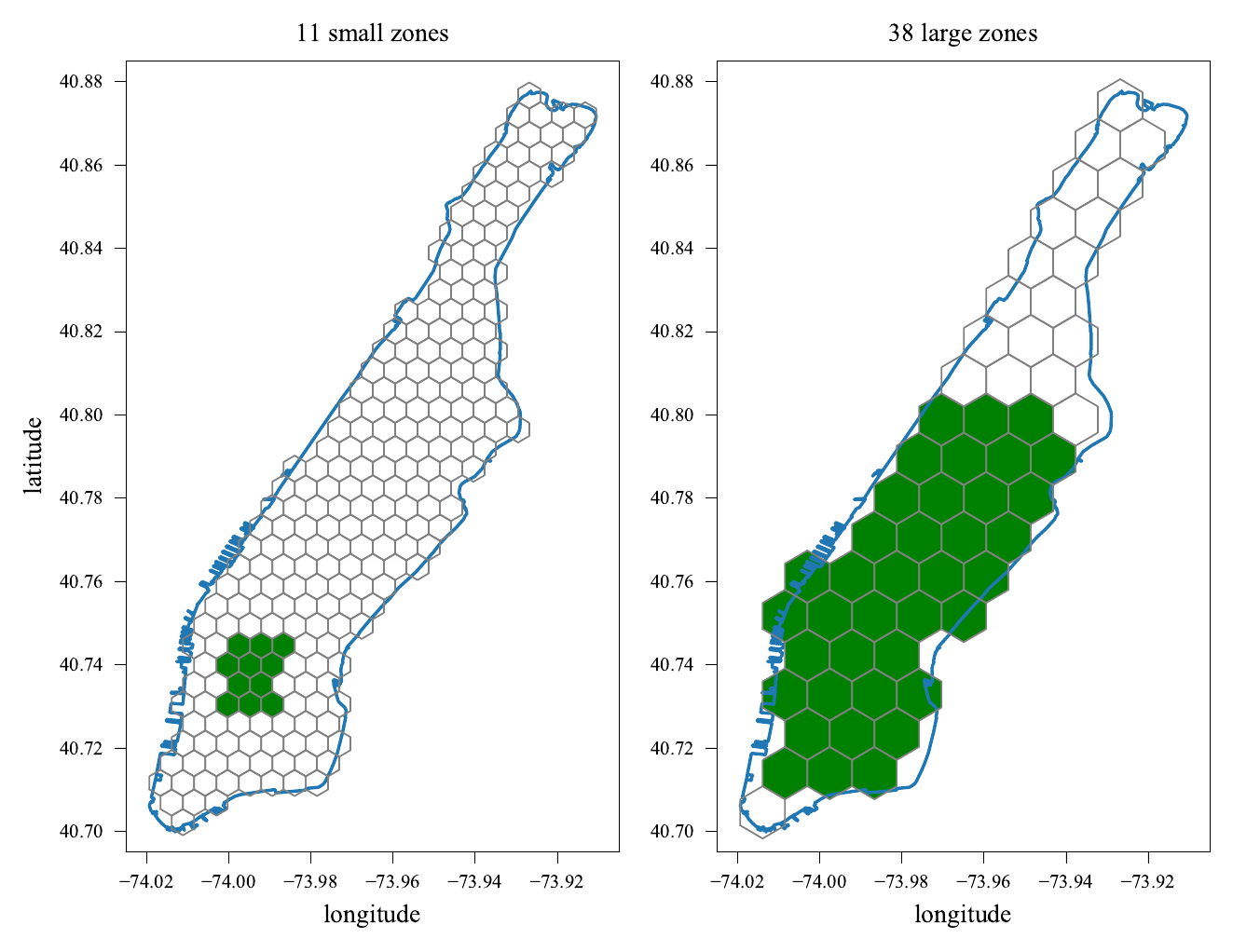}
    \caption{Hexagon grid laid over Manhattan for spatial discretization. The operating areas which we consider are marked in green.}
    \label{fig:hexagons}
\end{figure}

We use yellow taxi trip records from the year 2015 and exclude weekends and holidays. We assume that requests are placed at the time reported as the pickup time in the data set. Besides, we use the pickup and dropoff longitude/latitude from the data set and keep only trips for which pickup and dropoff coordinates are located on the main island of Manhattan. We discretize space with a hexagon grid as shown in Figure~\ref{fig:hexagons}. We assign a pickup and dropoff zone to each request based on the shortest distance from the longitude/latitude coordinates and remove trips that start and end in the same zone. 

Each zone is represented by a node in the graph $G$, that contains edges only between nodes that represent neighboring zones. The distance between neighboring zones is 459 meters and 917 meters for the small and large zones, respectively, and we assume a travel time of two and five time steps based on a realistic average driving speed. When vehicles travel between non-neighboring zones, they take the shortest route on $G$. 

We consider the two operating areas depicted in Figure~\ref{fig:hexagons}, i.e., we consider only requests that have a pickup and dropoff location within the green area. For the 38 large zones, we downscale the trip data by a factor of 20, i.e., we use only every 20\textsuperscript{th} request for our simulation, to have a system size suitable for the hardware that we used for the experiments. On average, this results in 360 requests per episode for the 11 small zones instance, with up to 23 requests in a single time step. For the 38 large zones instance, we observe 828 requests per episode on average, with up to 20 requests in a single time step. Note that the mean trip distance is larger for the 38 large zones instance, such that the number of vehicles required to serve a certain number of requests is larger than for 11 small zones. 

We assume a maximum waiting time of five minutes. To achieve an operating profit margin of 10\% when a request is served without empty driving to the pickup location, we set the revenue to 5.00 USD per km and the operational costs to 4.50 USD per km. Note that these numbers might be considered to be unrealistic, but can be scaled to a different level without any effect on the control problem and our results, since we report all results relative to the greedy performance. 

With the eight to 80 vehicles which we consider for the 11 small zones, the greedy policy serves 27\% to 78\% of the requests. For the 38 large zones with 50 to 250 vehicles, the greedy policy serves 30\% to 76\% of the requests. 

\subsection{Hyperparameters}
We train for 200,000 steps, update the network parameters every 20 steps, and test the performance of the current policy on the validation data every 2,880 steps (48 episodes). During the first 20,000 steps, we collect experience with a random policy and do not update the network parameters. 

For the critic loss, we use the Huber loss with a delta value of 10 instead of the squared error. Moreover, we use gradient clipping with a clipping ratio of 10 for actor and critic gradients. We use the Adam optimizer with a learning rate of $3\cdot10^{-4}$. We sample batches of size 128 from a replay buffer with maximum size 100,000. We set the discount factor to 0.9 since this gives a better and particularly more stable performance than other values which we tested. For the update of the target critic parameters we use an exponential moving average with smoothing factor $5\cdot10^{-3}$. We tune the entropy coefficient individually per instance and use values between 0.35 and 1.30 across the experiments reported in this paper. 

We repeat each training run with multiple random seeds and use the model with the best validation performance across runs to test the performance of our method on the test data set. Results reported throughout this paper correspond to these test results. For the 11 small zones instance, we use five random seeds, while we use three random seeds for the 38 large zones instance. 

The MPC results are based on an average over multiple random seeds. For the results in Figure~\ref{fig:box_plots}, we use five random seeds, while we use three random seeds for the results in Figure~\ref{fig:kl_divergence}.

\subsection{Benchmarks}
We benchmark our algorithm against two well-known policies that are as follows.

\textbf{Greedy.} The greedy policy considers requests in their arrival order. Whenever there is at least one vehicle that will be able to serve the request within the maximum waiting time and with a positive profit, the greedy policy accepts the request. If there is no such vehicle, the policy rejects the request. The profit calculation takes into account the revenue from serving the request and the cost to drive from the request's origin to its destination, as well as the cost to drive from the destination of the request that the vehicle currently serves (the position of the vehicle if it is idle) to the origin of the new request. If there is more than one vehicle which fulfills these conditions, the greedy policy assigns the request to the vehicle that will be closest to the request origin once it has finished its current job. 

\textbf{MPC.} We adapt the approach by \cite{AlonsoMora2017} to our problem setting. First, for each 15 minute interval, we estimate a probability distribution over origin-destination pairs from the training data, i.e., we obtain the probability that a new request shall be picked up at this origin and dropped off at this destination based on frequentist statistics. Here, we use Laplace smoothing to mitigate a bias from potentially sparse data. In addition, we estimate the number of requests that can be expected. We then use those estimates for online decision-making as follows: in each time step, we observe the new (real) requests. In addition, for some sampling horizon, we sample the expected number of requests from the estimated probability distribution (virtual requests). With the current vehicle states, real and virtual requests, we solve an offline optimization problem, maximizing for total profit assuming perfect information over the sampling horizon, with mixed integer programming. The solution gives a decision for the real requests, which we use as the action for the current time step. We repeat this process in a receding horizon fashion. We set the sampling horizon to five minutes, which turned out to yield the best performance.

\section{Complementary Results}
\begin{figure}
    \centering
    \includegraphics[width=\textwidth]{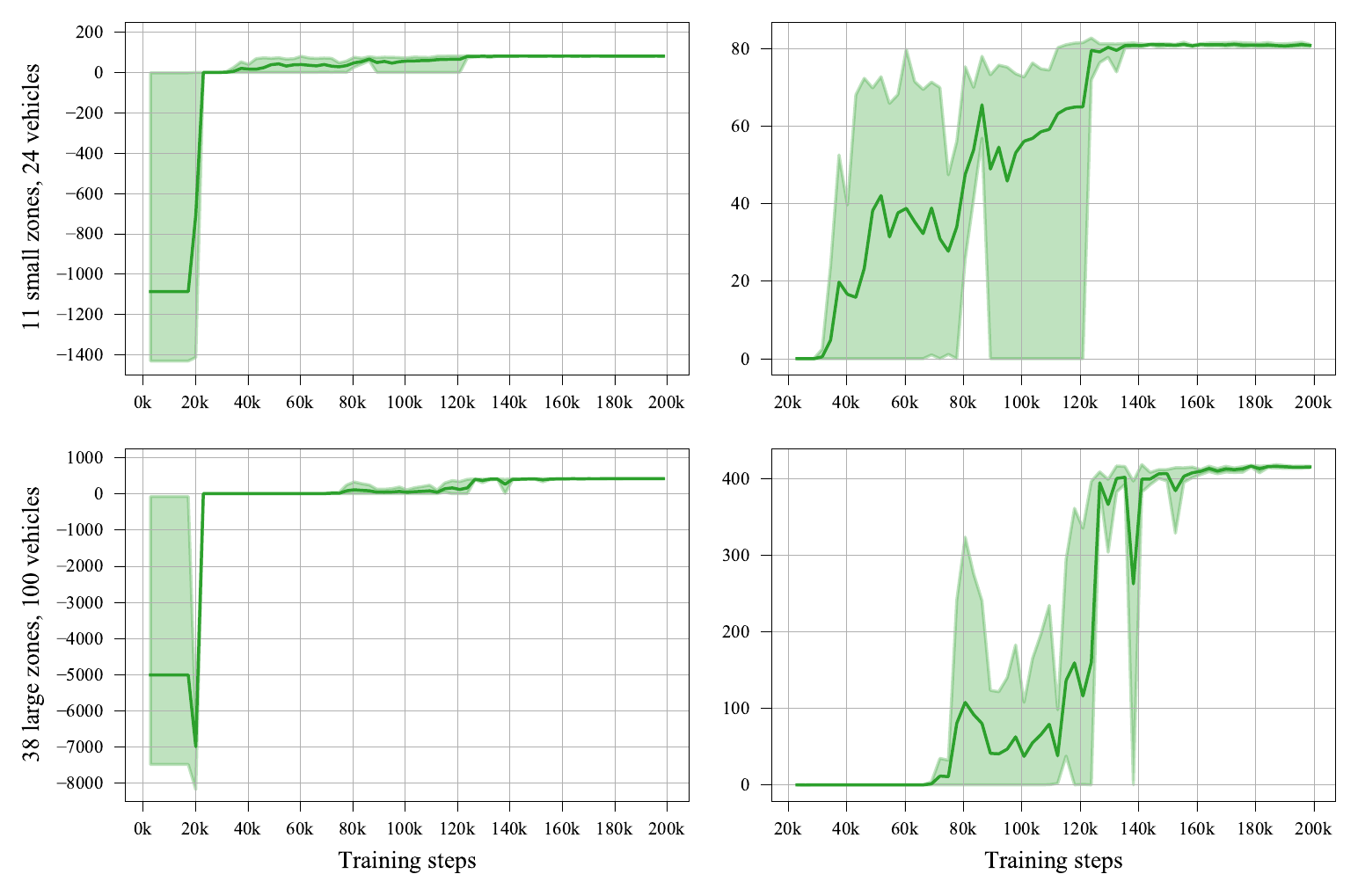}
    \caption{Validation reward along the training process for two exemplary instances. The green line is the mean over random seeds, the shaded area depicts the minimum and maximum values over random seeds. Plots in the right column are zoomed in versions of the plots in the left column to make the part of the training process after a reward of zero is reached better visible.}
    \label{fig:validation_reward}
\end{figure}

\begin{figure}
    \centering
    \includegraphics{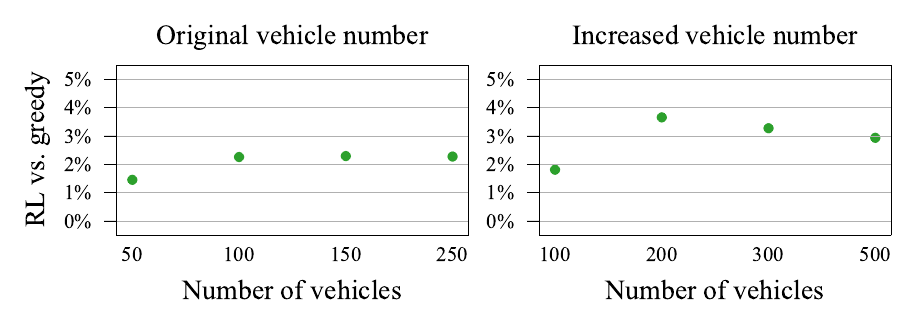}
    \caption{Average performance of our method vs. greedy over the 20 test dates for the 38 large zones instance. We train the RL agents with 50, 100, 150, and 250 vehicles on the original system. The left plot shows the test performance on the original system. The right plot shows the test performance of the same policies, without additional training, for an increased system size, with twice as many vehicles and requests (downscaled by factor 10 instead of 20), i.e., we use the policy trained with 50 vehicles for the 100 vehicles on the larger system, and likewise for the other vehicle numbers.}
    \label{fig:different_agent_count}
\end{figure}

In the following, we provide additional result plots. Figure~\ref{fig:validation_reward} shows the validation reward to illustrate the training process for two exemplary instances: 11 small zones with 24 vehicles and 38 large zones with 100 vehicles. Figure~\ref{fig:different_agent_count} shows the test performance of a policy trained with some original number of agents in a system with an increased number of agents. The performance does not deteriorate, illustrating that our method is non-parametric w.r.t. the number of agents.

\newpage

% Acknowledgments---Will not appear in anonymized version
\acks{We thank Gerhard Hiermann for his help with the implementation of the MIP for the MPC-based benchmark. The work of Tobias Enders was supported by a fellowship within the IFI program of the German Academic Exchange Service (DAAD). This work was supported by the German Research Foundation (DFG) under grant no. 449261765.}

\bibliography{bibfile}

\end{document}